\documentclass[runningheads]{llncs}
\usepackage[T1]{fontenc}
\usepackage{latexsym}
\usepackage{amssymb}
\usepackage{amsmath}
\usepackage{booktabs}
\usepackage{enumitem}
\usepackage{color}
\usepackage{multirow} 
\usepackage{algorithm}
\usepackage{algorithmic}
\usepackage{subcaption}
\usepackage{xcolor}
\usepackage{array}
\usepackage{amssymb,amsfonts}
\usepackage{enumitem}
\usepackage{graphicx}
\begin{document}
\title{Spatial-ViLT: Enhancing Visual Spatial Reasoning through Multi-Task Learning}
\titlerunning{Spatial-ViLT}
%
%

\author{%
  Chashi Mahiul Islam\inst{1}\orcidID{0009-0000-4048-5784}\textsuperscript{*} \and
  Oteo Mamo\inst{1}\orcidID{0000-0003-0865-0019}\textsuperscript{*} \and
  Samuel Jacob Chacko\inst{1}\orcidID{0009-0006-0948-4234} \and
  Xiuwen Liu\inst{1}\orcidID{0000-0002-9320-3872} \and
  Weikuan Yu\inst{1}\orcidID{0000-0002-8754-0311}
}

\authorrunning{Islam et al.}
\institute{Florida State University, Department of Computer Science, Tallahassee, Florida, USA\\
\email{\{ci20l, om21d, sj21j, liux, wyu3\}@fsu.edu}}

\maketitle

\begingroup
  \renewcommand\thefootnote{\fnsymbol{footnote}}%
  \footnotetext[1]{These authors contributed equally.}%
\endgroup

%
\begin{abstract}
Vision-language models (VLMs) have advanced multimodal reasoning but still face challenges in spatial reasoning for 3D scenes and complex object configurations. To address this, we introduce SpatialViLT, an enhanced VLM that integrates spatial features like depth maps, 3D coordinates, and edge maps through a multi-task learning framework. This approach enriches multimodal embeddings with spatial understanding. We propose two variants: SpatialViLT and MaskedSpatialViLT, focusing on full and masked object regions, respectively. Additionally, SpatialEnsemble combines both approaches, achieving state-of-the-art accuracy. Our models excel in spatial reasoning categories such as directional, topological, and proximity relations, as demonstrated on the challenging Visual Spatial Reasoning (VSR) dataset. This work represents a significant step in enhancing the spatial intelligence of AI systems, crucial for advanced multimodal understanding and real-world applications.

\keywords{Vision-Language Models, Spatial Reasoning, Multi-task Learning, 3D Scene Understanding, Visual-Spatial Features}
\end{abstract}

\section{Introduction}
\label{sec:intro}

\begin{figure}[ht]
    \centering
    \includegraphics[width=0.9\textwidth]{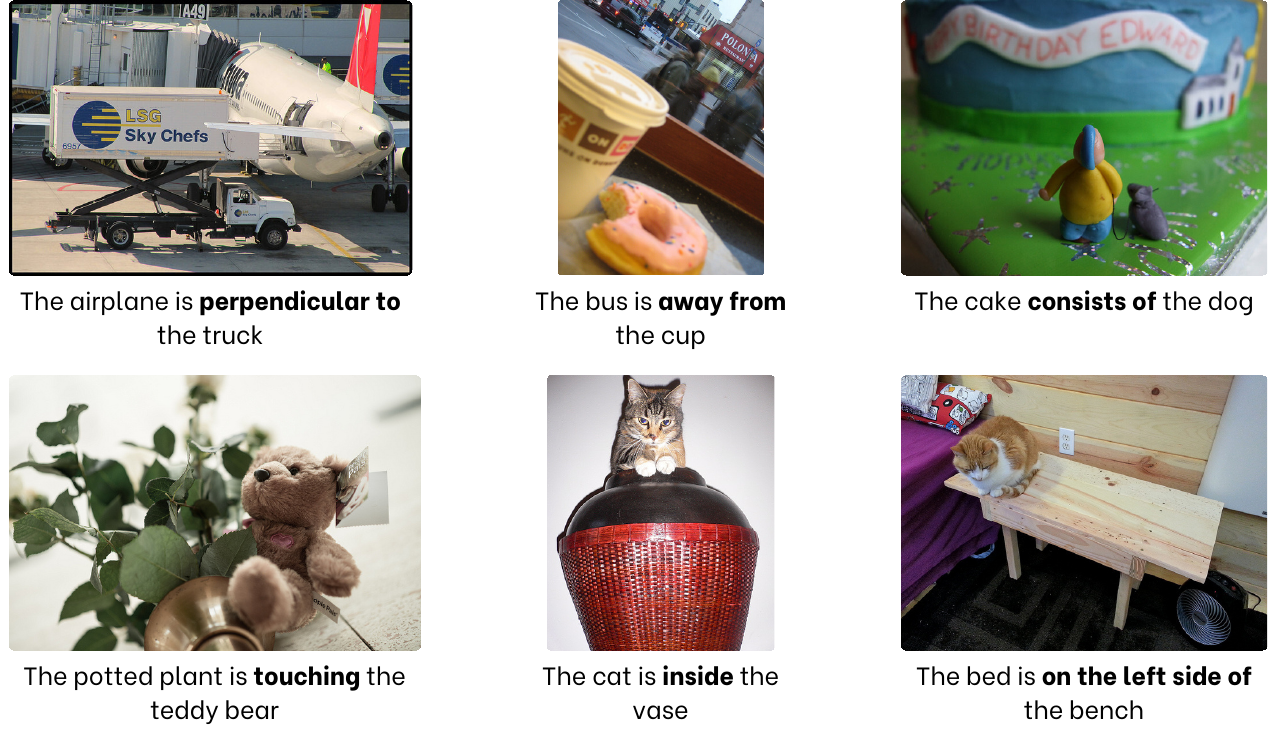}
    \caption{Challenging spatial reasoning examples from the VSR dataset that highlight current limitations in vision-language models. These cases demonstrate failures in orientation understanding (airplane-truck), proximity detection in complex scenes (bus-cup), topological reasoning (cat-vase), semantic consistency (cake-dog), contact detection (plant-teddy bear), and precise spatial positioning (bed-bench).}
    \label{fig:motivating_examples}
\end{figure}

Vision-language models (VLMs) have achieved impressive performance on various multimodal reasoning tasks like visual question answering, image captioning, and multimodal
 verification \cite{long_hoang_dang__2024}. However, their ability to comprehend and reason about spatial relationships between objects, a core aspect of human cognition, remains limited \cite{fangjun_li__2024}. Spatial reasoning is critical for AI systems to develop a deeper understanding of the spatial configurations and interactions in the 3D world around us \cite{chen2024spatialvlm}. Figure \ref{fig:motivating_examples} illustrates the types of spatial reasoning challenges that current VLMs struggle with, spanning multiple categories of spatial relations from orientation and proximity to topological and semantic understanding. 

Recent work like SpatialVLM \cite{chen2024spatialvlm} and \cite{Rajabi2023TowardsGV} has attempted to address this shortcoming by creating large-scale datasets focused on spatial reasoning, and fine-tuning VLMs explicitly on spatial relation understanding tasks respectively. However, even models fine-tuned on these spatial reasoning datasets struggle to achieve human-level performance on established benchmarks like the VSR dataset ~\cite{liu2023visual}. Prior spatial reasoning tools have leveraged various factors such as explicit location information \cite{azkune2024grounding}, top-view perspectives \cite{li2024topviewrs}, and depth information \cite{cheng2024spatialrgpt} to enhance model performance. Despite these advancements, significant deficiencies remain, particularly in the attention given to object shapes and characteristics. Many models still lack robust spatial priors and 3D understanding in visual encoding, leading to performance gaps compared to human-level reasoning \cite{liu2023visual}. This highlights the ongoing need for improved integration of these features to achieve more accurate and comprehensive spatial reasoning in vision-language models.
 
\begin{figure*}[ht]
    \centering
    \includegraphics[width=.9\textwidth]{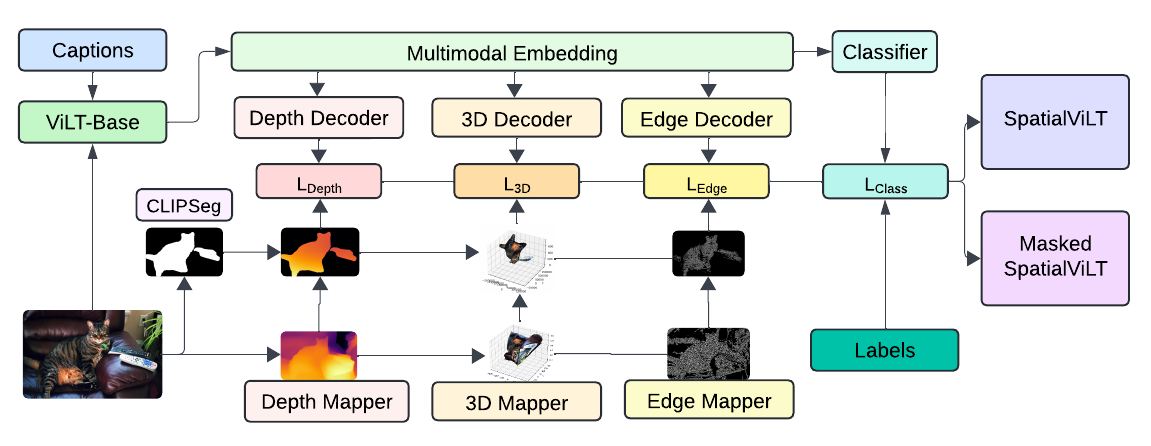}
\caption{Framework for training SpatialViLT and Masked SpatialViLT. Captions and images are processed using ViLT (Base), with depth, 3D, and edge features extracted and decoded to calculate corresponding losses. These losses are backpropagated to improve the multimodal embeddings with spatial priors for enhanced spatial reasoning and classification.}
    \label{fig:spatialvilt}
\end{figure*}

We investigate vision-language models' limitations in spatial reasoning using the VSR dataset, which categorizes spatial relations into meta-categories. Identifying the need for spatial priors and 3D understanding, we propose SpatialViLT and its masked variant. These enhanced vision-language models are trained via multi-task learning (Figure \ref{fig:spatialvilt}) to predict spatial representations like depth maps, 3D coordinate maps, and edge maps from multi-modal embeddings, improving spatial understanding.

To address different spatial relation meta-categories, we create SpatialEnsemble, a relation-aware branching ensemble combining predictions from multiple customized SpatialViLT models along with other models. This architecture achieves state-of-the-art performance on the VSR benchmark. Our object-level analysis reveals that models struggle with spatial relations involving animate objects with distinguishable frontal surfaces or faces. To the best of our knowledge, we are the first to combine these spatial techniques in a multi-task learning framework for vision-language models, achieving state-of-the-art performance in visual spatial reasoning tasks.
Our key contributions are:
\begin{itemize}[left=0.1cm,nosep]
\item SpatialViLT and MaskedSpatialViLT, two enhanced vision-language models trained via a multi-modal multitask learning framework for richer spatial priors and 3D understanding.
\item SpatialEnsemble, a relation-aware branching ensemble combining specialized spatial relation experts, achieving state-of-the-art spatial reasoning performance.
\end{itemize}

\section{Related Work}
\label{sec:related}

Visual spatial reasoning, the ability to understand spatial relationships between objects in visual scenes, is a crucial yet challenging problem in computer vision and natural language processing. Recent work has focused on developing both specialized datasets and improved architectures to address this challenge.

The VSR dataset \cite{liu2023visual} stands out among spatial reasoning datasets, containing over 10,000 text-image pairs annotated with 66 types of spatial relations. Its complexity stems from challenging linguistic phenomena like varying reference frames, making it a robust benchmark for evaluating spatial reasoning capabilities. Recent datasets have further enriched the field by integrating 3D geometric information \cite{3d_vqa} and exploring spatial perspective-taking in embodied environments \cite{embodied}.

Different approaches have been proposed to tackle visual spatial reasoning. Text-only language models have shown promise in grounding spatial relations using explicit location information \cite{balazevic2024towards}, while approaches like SpatialRGPT \cite{cheng2024spatialrgpt} have enhanced spatial perception by integrating depth information and 3D scene graphs. State-of-the-art vision-language models including LXMERT \cite{tan2019lxmert}, ViLT \cite{kim2021vilt}, and VisualBERT \cite{li2019visualbert} have advanced VSR tasks by effectively combining visual and textual information. However, studies have highlighted continuing challenges, particularly in handling spatial prepositions \cite{kamath2023s} and object localization \cite{Rajabi2023TowardsGV}. Our work builds on these foundations through a multi-task learning paradigm \cite{singh2022design} that integrates depth cues, edge maps, and 3D spatial configurations, moving towards bridging the gap between current model performance and human-level spatial reasoning.

\section{Preliminaries}
\label{sec:prelim}

In this section, we discuss the foundational elements of our study, including the dataset and model selection. We chose the VSR dataset for its extensive range of spatial relations and complex scenarios, making it an ideal benchmark for evaluating the spatial reasoning capabilities of vision-and-language models (VLMs). For our experiments, we utilize the Vision-and-Language Transformer (ViLT) due to its efficient architecture and suitability for integrating various spatial features such as depth, edge, and 3D coordinate information \cite{kim2021vilt}.

\subsection{VSR Dataset: Meta-Categories and Corresponding Features}
The VSR dataset \cite{liu2023visual} evaluates VLMs' spatial reasoning abilities with over 10,000 text-image pairs annotated for 71 spatial relations across seven meta-categories. Images from the MS COCO 2017 dataset were annotated to emphasize spatial relations and validated to ensure quality. Despite a human accuracy ceiling of 95.4\%, current VLMs like VisualBERT, LXMERT, ViLT, and CLIP only achieve no more than 70\%, showcasing the dataset's difficulty and the gap in model capabilities.

\begin{table}[ht]
\centering
\resizebox{\columnwidth}{!}{%
\begin{tabular}{|p{2cm}|p{6cm}|p{4cm}|}
\hline
\textbf{Meta-Category} & \textbf{Relations} & \textbf{Spatial Features Required} \\ \hline

Adjacency &
adjacent to, alongside, at the side of, at the right side of, at the left side of, attached to, at the back of, ahead of, against, at the edge of &
Edge detection, Surface Normals, Depth maps, 3D coordinates \\ \hline

Directional &
off, past, toward, down, deep down, up, away from, along, around, from, into, to, across, across from, through, down from &
Depth maps, 3D coordinates, Motion vectors, Trajectory analysis \\ \hline

Orientation &
facing, facing away from, parallel to, perpendicular to &
 Pose estimation, 3D coordinates, Edge maps \\ \hline

Projective &
on top of, beneath, beside, behind, left of, right of, under, in front of, below, above, over, in the middle of &
Relative Depth Estimation, Depth maps, Occlusion Detection \\ \hline

Proximity &
by, close to, near, far from, far away from &
Depth maps, Distance measurement \\ \hline

Topological &
connected to, detached from, has as a part, part of, contains, within, at, on, in, with, surrounding, among, consists of, out of, between, inside, outside, touching &
Segmentation, Depth maps, 3D coordinates, Edge maps \\ \hline

Unallocated &
beyond, next to, opposite to, after, enclosed by &
Depth maps, 3D coordinates, Relative positioning \\ \hline

\end{tabular}
}
\caption{The table categorizes spatial relations from the VSR Dataset into meta-categories and identifies the essential spatial features necessary for accurate interpretation and reasoning in vision-and-language models.}
\label{table:spatial_features}
\end{table}

Table \ref{table:spatial_features} presents the meta-categories and their associated relations used in the VSR dataset. Adapted from the summarized table in \cite{marchi2021cross}, we have included the spatial features required for each meta-category based on detailed study. This table categorizes spatial relations and identifies essential spatial features necessary for accurate interpretation. Research highlights the importance of these features: depth maps are crucial for understanding adjacency and projective relations \cite{ranftl2020towards}, 3D coordinates facilitate orientation and spatial positioning \cite{qi2017pointnet}, and edge detection is vital for identifying boundaries and topological relationships \cite{canny1986computational}. Directional relations benefit from motion vectors and trajectory analysis \cite{yi2019clevrer}, while pose estimation and orientation detection are critical for accurately interpreting orientation \cite{chen2020monocular}. Distance measurement tools enhance proximity understanding \cite{dai2017scannet}, and segmentation aids in recognizing topological relations \cite{he2017mask}. These insights underscore the necessity of integrating these spatial features into VLMs to enhance their performance in spatial reasoning tasks.

\subsection{ViLT Model}
The Vision-and-Language Transformer (ViLT) \cite{kim2021vilt} integrates visual and textual information using a transformer model. It tokenizes text and divides images into patches projected into the same embedding space. These embeddings are processed through a transformer encoder, with the pooled [CLS] token representation used for downstream tasks. ViLT's efficiency and adaptability make it suitable for multitask learning involving depth, edge, and 3D coordinate masks, despite LXMERT's state-of-the-art accuracy.

\section{Methodology}
\label{sec:method}

In this section, we present our comprehensive approach to enhancing the multimodal embeddings of the ViLT model with spatial priors. Our methodology focuses on predicting additional spatial features (depth, 3D coordinates, and edge information) in a multitask learning framework. We describe the processes of object segmentation, depth map creation, 3D coordinates extraction, and edge map generation, as well as the different model variants and ensemble techniques employed to improve spatial reasoning capabilities.

\subsection{SpatialViLT: Integrating Spatial Information}

Our primary objective is to enhance the multimodal embedding of the ViLT model with spatial priors through two distinct model variants. The first variant, SpatialViLT, integrates spatial features (depth, 3D coordinates, edge) into the ViLT model without applying object masks, aiming to capture global spatial information across the entire image. The second variant, MaskedSpatialViLT, applies these spatial features with object masks, enhancing focus on relevant regions and allowing the model to concentrate on object-specific spatial relationships.

\subsubsection{Feature Extraction Pipeline}

\textbf{Object Segmentation}
We employ the CLIPSeg model to create masks for the objects in the image. The CLIPSeg model generates high-quality masks based on textual descriptions \cite{luddecke2022image}, with masks being resized and applied to isolate objects of interest.

\textbf{Depth Map Creation}
Using a pre-trained depth estimation model from the MiDaS framework \cite{ranftl2020towards}, we generate depth maps where each pixel value represents the distance from the camera.

\textbf{3D Coordinates Extraction}
We derive 3D coordinates using the equations $x = (u - c_x) \cdot z / f_x$ and $y = (v - c_y) \cdot z / f_y$, where $u$ and $v$ are pixel coordinates, $c_x$ and $c_y$ are camera center coordinates, $f_x$ and $f_y$ are focal lengths (assumed unit value), and $z$ is the depth value \cite{zhou2018open3d}.

\textbf{Edge Map Generation}
We apply the Canny edge detection algorithm \cite{4767851} to grayscale-converted images, producing binary edge maps that highlight object boundaries.

\begin{figure*}[ht]
    \centering
    \includegraphics[width=1\textwidth]{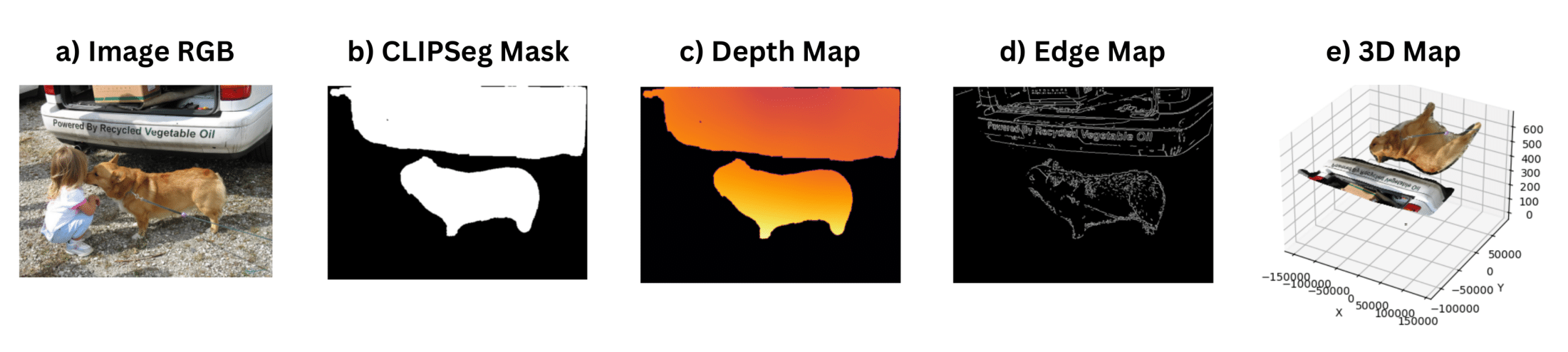}
\caption{Feature extraction pipeline demonstrating the multi-modal spatial feature generation process: (a) Original RGB image, (b) CLIPSeg-generated object masks for "dog" and "truck" extracted from caption text, (c) MiDaS depth map showing relative distances, (d) Canny edge map highlighting object boundaries, and (e) 3D coordinate map derived from depth information.}
    \label{fig:feature_extraction}
\end{figure*}

Figure \ref{fig:feature_extraction} illustrates our comprehensive feature extraction pipeline that transforms RGB images into multiple spatial representations, enabling our model to learn rich spatial priors for enhanced reasoning capabilities.

\subsubsection{Training Framework}

The key components of our architecture include a pre-trained ViltModel, CNN encoders for spatial feature processing, corresponding decoders for reconstruction, and a classifier for final predictions. The training process is formalized in Algorithm \ref{alg:spatialvilt_training}, which details our multitask learning approach with weighted loss components.

\begin{algorithm}[t!]
\caption{SpatialViLT Training}
\label{alg:spatialvilt_training}
\begin{algorithmic}[1]
\REQUIRE Image set $I$, caption set $C$, label set $L$
\ENSURE Trained SpatialViLT model $\mathcal{M}$

\STATE $\mathcal{M} \gets$ InitializeSpatialViLTModel() 

\WHILE{not converged}
    \STATE $(I_b, C_b, L_b) \gets$ SampleBatch($I, C, L$) 
    \STATE $X \gets$ ProcessInputs($I_b, C_b$) 
    \STATE $D \gets$ ExtractDepthMaps($I_b$) 
    \STATE $R \gets$ Compute3DCoordinates($D$) 
    \STATE $E \gets$ ExtractEdgeMaps($I_b$) 
    
    \STATE $(y, \hat{D}, \hat{R}, \hat{E}) \gets \mathcal{M}(X, D, R, E)$ \COMMENT{Forward pass through model: predicts class labels $y$ and reconstructs spatial features ($\hat{D}$ depth, $\hat{R}$ 3D coordinates, $\hat{E}$ edge maps)}
    
    \STATE $\mathcal{L}_c \gets$ ComputeClassificationLoss($y, L_b$) 
    \STATE $\mathcal{L}_d \gets$ ComputeDepthReconstructionLoss($\hat{D}, D$) 
    \STATE $\mathcal{L}_r \gets$ Compute3DReconstructionLoss($\hat{R}, R$) 
    \STATE $\mathcal{L}_e \gets$ ComputeEdgeReconstructionLoss($\hat{E}, E$) 
    
    \STATE $\mathcal{L}_{\text{total}} \gets \mathcal{L}_c + \lambda_d\mathcal{L}_d + \lambda_r\mathcal{L}_r + \lambda_e\mathcal{L}_e$ 
    
    \STATE UpdateModel($\mathcal{M}, \mathcal{L}_{\text{total}}$) 
\ENDWHILE

\STATE \textbf{return} $\mathcal{M}$
\end{algorithmic}
\end{algorithm}

\subsection{SpatialEnsemble Technique}

The SpatialEnsemble model employs a performance-based weighted voting technique \cite{li2020performance}, leveraging the strengths of multiple models: LXMERT, ViLT, SpatialViLT, and MaskedSpatialViLT. The weight for each model is determined by its accuracy on a validation set:

\begin{equation}
    w_{i,c,r} = \frac{a_{i,c,r}}{\sum_{j=1}^{N} a_{j,c,r}}
\end{equation}

where $w_{i,c,r}$ represents the weight for model $i$, meta-category $c$, and relation $r$. For each instance in the test set, the ensemble computes:

\begin{equation}
    s_c = \sum_{i=1}^{N} w_{i,c,r} \cdot \mathbb{I}(p_i = c)
\end{equation}

where $s_c$ is the score for class $c$, $p_i$ is the prediction of model $i$, and $\mathbb{I}$ is the indicator function.
\section{Experimental Results and Discussion}

\subsection{Experimental Setup}
We evaluate our models on the Visual Spatial Reasoning (VSR) dataset, comprising 10,972 image-caption pairs. The dataset was split into 70\% training (7,678 pairs), 10\% validation (1,098 pairs), and 20\% testing (2,196 pairs). We trained LXMERT and ViLT for 20 epochs (lr=2e-5), and our SpatialViLT variants for 15 epochs (lr=1e-4), implementing early stopping to prevent overfitting.

We compare our specialized models (SpatialViLT, MaskedSpatialViLT, and SpatialEnsemble) against state-of-the-art baselines: LXMERT, ViLT \cite{liu2023visual}, and SpaceLLaVA \cite{chen2024spatialvlm}. Our evaluation focuses on model performance across seven meta-categories of spatial relationships.

\begin{table*}[ht]
    \caption{Average performance comparison of ViLT variants across meta-categories (in \%). Bold indicates best performance for each category.}
    \label{tab:model_comparison_focused}
    \centering
    \resizebox{\textwidth}{!}{%
    \begin{tabular}{|l|c|c|c|c|c|c|}
        \hline
        \textbf{Meta-Category} & \textbf{LXMERT (\%)} & \textbf{ViLT (\%)} & \textbf{SpaceLLaVa (\%)} & \textbf{MaskedSpatialViLT (\%)} & \textbf{SpatialViLT (\%)} & \textbf{Best Model} \\
        \hline
        Proximity    & 73.44 & 76.56 & 73.47 & 69.53 & \textbf{77.08} & SpatialViLT \\
        Unallocated  & 67.11 & 64.47 & 50.00 & 57.89 & \textbf{72.73} & SpatialViLT \\
        Orientation  & 53.28 & 57.66 & 55.93 & 54.74 & \textbf{61.02} & SpatialViLT \\
        Topological  & 72.80 & 75.08 & 75.24 & \textbf{75.90} & 73.87 & MaskedSpatialViLT \\
        Directional  & 66.67 & 64.81 & 60.98 & \textbf{68.52} & 63.46 & MaskedSpatialViLT \\
        Adjacency    & \textbf{64.71} & 58.13 & 58.60 & 59.86 & 60.75 & LXMERT \\
        Projective   & \textbf{72.36} & 67.85 & 66.83 & 63.23 & 66.08 & LXMERT \\
        \hline
    \end{tabular}
    }
\end{table*}

\subsection{Individual Model Performance Analysis}
As shown in Table \ref{tab:model_comparison_focused}, SpatialViLT demonstrated superior performance in several categories by leveraging global spatial features effectively. It achieved the highest accuracy in proximity (77.08\%), orientation (61.02\%), and unallocated (72.73\%) categories. MaskedSpatialViLT excelled in different areas, performing best in topological (75.90\%) and directional (68.52\%) categories through its focused object-specific spatial features. The complementary strengths of both variants suggest that our spatial features successfully approximate additional spatial cues.

\begin{table*}[ht]
    \caption{SpatialEnsemble average performance comparison across meta-categories (in \%). Bold indicates the best performance for each category. Improvement shows SpatialEnsemble's gain over the best baseline.}
    \label{tab:meta_category_comparison}
    \centering
    \resizebox{\textwidth}{!}{%
    \begin{tabular}{|l|c|c|c|c|c|}
        \hline
        \textbf{Meta-Category} & \textbf{LXMERT (\%)} & \textbf{ViLT (\%)} & \textbf{SpaceLLaVa (\%)} & \textbf{SpatialEnsemble (\%)} & \textbf{Improvement (\%)} \\
        \hline
        Unallocated  & 67.11 & 64.47 & 50.00 & \textbf{72.73} & +5.62 \\
        Directional  & 66.67 & 64.81 & 60.98 & \textbf{72.22} & +5.55 \\
        Topological  & 72.80 & 75.08 & 75.24 & \textbf{79.80} & +4.72 \\
        Proximity    & 73.44 & 76.56 & 73.47 & \textbf{79.69} & +3.13 \\
        Adjacency    & 64.71 & 58.13 & 58.60 & \textbf{65.74} & +1.03 \\
        Projective   & \textbf{72.36} & 67.85 & 66.83 & \textbf{72.36} & 0.00 \\
        Orientation  & 53.28 & \textbf{57.66} & 55.93 & 54.74 & -2.92 \\
        \hline
        \textbf{Overall} & 69.88 & 69.84 & 65.70 & \textbf{72.62} & +2.74 \\
        \hline
    \end{tabular}
    }
\end{table*}

\subsection{SpatialEnsemble Performance Analysis}
The SpatialEnsemble model achieved significant improvements across most meta-categories (Table \ref{tab:meta_category_comparison}). Notable gains were observed in unallocated (+5.62\%), directional (+5.55\%), and topological (+4.72\%) categories. The ensemble maintained consistent performance in the projective category, matching LXMERT's accuracy of 72.36\%.

\begin{figure}[t]
    \centering
    \includegraphics[width=0.98\textwidth]{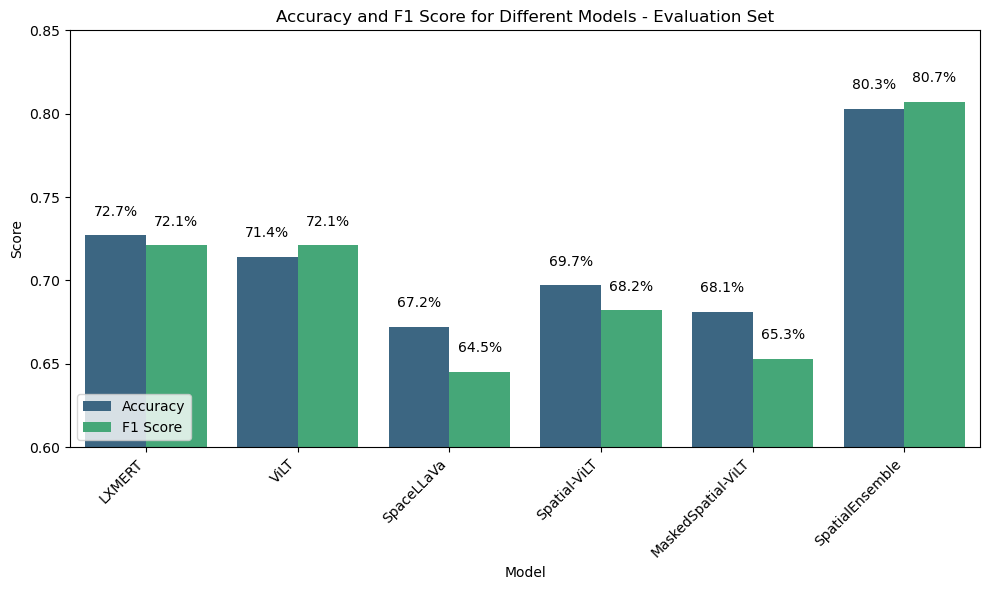}
    \caption{Accuracy and F1 Score of Different Models on the Evaluation Set.}
    \label{fig:eval_models}
\end{figure}

\begin{figure}[t!]
    \centering
    \includegraphics[width=0.98\textwidth]{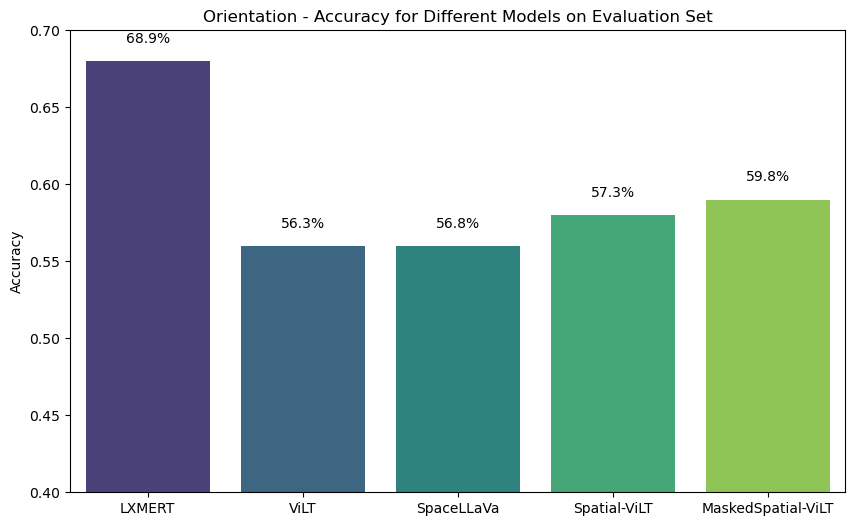}
    \caption{Accuracy of Different Models for Orientation Meta-Category on the Evaluation Set.}
    \label{fig:orientation_eval}
\end{figure}

However, we observed an unexpected challenge in the orientation category, where performance decreased by 2.92\% compared to the ViLT baseline. This decline was traced to LXMERT's inconsistent performance between evaluation and test sets, as illustrated in Figure \ref{fig:orientation_eval}. While LXMERT showed superior performance (68\%) during evaluation, leading to higher weights in the ensemble, its significant performance drop on the test set adversely affected the overall ensemble predictions for orientation-related tasks.

\subsection{Overall Performance and Future Directions}
SpatialEnsemble achieved state-of-the-art performance with 72.62\% overall accuracy and 72.64\% F1 score, surpassing both LXMERT (69.88\%) and SpaceLLaVA (65.70\%). However, we identified a significant generalization gap between evaluation and test performance. While showing an 8\% improvement on the evaluation set, this advantage reduced to 2.74\% on the test set (Figure \ref{fig:eval_models}). This generalization gap could be addressed in future work through dynamic weight adjustment mechanisms and cross-validation techniques to obtain more robust ensemble predictions across different data distributions. Further analysis revealed a performance disparity between animate and inanimate objects in orientation and projection tasks.

This finding suggests that incorporating pose estimation features could potentially improve model performance for animate objects, pointing to a promising direction for future work.

\section{Conclusion}

In this paper, we introduced SpatialViLT and MaskedSpatialViLT, two novel multi-modal multitask learning frameworks that enhance spatial reasoning in vision-language models. By incorporating auxiliary spatial prediction tasks such as depth map, edge map, and 3D coordinate reconstruction, these models enrich their visual encoders' spatial awareness in distinct ways. Our evaluations demonstrate that SpatialViLT and its ensemble variant SpatialEnsemble (which combines SpatialViLT variants with LXMERT) significantly outperform the state-of-the-art LXMERT model, particularly in handling complex spatial relations involving directional, topological, proximity, and unallocated meta-categories. Future enhancements could include pose estimation and trajectory analysis to improve the models' ability to distinguish orientation between objects. Our proposed frameworks pave the way for more robust spatial reasoning in vision-language tasks.

\section*{Acknowledgments}
This work is supported in part by the National Science Foundation awards 1763547 and 2403089, and has used the AWS services funded by a grant by the Florida State University and the NoleLand facility that is funded by the U.S. National Science Foundation grant CNS-1822737.
\bibliographystyle{splncs04}
\bibliography{main}

\end{document}